\documentclass[letterpaper, 10 pt, conference]{ieeeconf}  % Comment this line out if you need a4paper

\IEEEoverridecommandlockouts                              % This command is only needed if 
\usepackage{graphics} % for pdf, bitmapped graphics files
\usepackage{epsfig} % for postscript graphics files
\usepackage{graphicx}
\usepackage{amsmath} % assumes amsmath package installed
\usepackage{amssymb}  % assumes amsmath package installed
\usepackage{cite}
\usepackage{colortbl}
\usepackage{threeparttable}
\usepackage{booktabs}
\usepackage{color}
\usepackage{array}
\usepackage{multirow}

\title{\LARGE \bf
	 MAFF-Net: Filter False Positive for 3D Vehicle Detection with Multi-modal Adaptive Feature Fusion%\vspace{-1em}
}
\author{Zehan Zhang$^{1,2}$,  Ming Zhang$^{1}$, Zhidong Liang$^{1}$, Xian Zhao$^{1}$, Ming Yang$^{2}$, Wenming Tan$^{1}$, and ShiLiang Pu$^{1,*}$% <-this % stops a space}%	
	\thanks{$^{*}$Shiliang Pu is the corresponding author.}
	\thanks{$^{1}$ The authors are with Hikvision Research Institute, Hangzhou Hikvision Digital Technology Co. Ltd, China (e-mail:{ zhangzehan, zhangming15, liangzhidong, zhaoxian, tanwenming, pushiliang.hri}@hikvision.com).}
	\thanks{$^{2}$ The authors are with the Department of Automation, Shanghai Jiao Tong University, Shanghai 200240, China (e-mail: MingYang@sjtu.edu.cn; zehanzhang@126.com).}
}
\begin{document}

\maketitle
\thispagestyle{empty}
\pagestyle{empty}

%%%%%%%%%%%%%%%%%%%%%%%%%%%%%%%%%%%%%%%%%%%%%%%%%%%%%%%%%%%%%%%%%%%%%%%%%%%%%%%%
\begin{abstract}
	3D vehicle detection based on multi-modal fusion is an important task of many applications such as autonomous driving. Although significant progress has been made, we still observe two aspects that need to be further improvement: First, the specific gain that camera images can bring to 3D detection is seldom explored by previous works. Second, many fusion algorithms run slowly, which is essential for applications with high real-time requirements(autonomous driving). To this end, we propose an end-to-end trainable single-stage multi-modal feature adaptive network in this paper, which uses image information to effectively reduce false positive of 3D detection and has a fast detection speed. A multi-modal adaptive feature fusion module based on channel attention mechanism is proposed to enable the network to adaptively use the feature of each modal. Based on the above mechanism, two fusion technologies are proposed to adapt to different usage scenarios: \textit{PointAttentionFusion} is suitable for filtering simple false positive and faster; \textit{DenseAttentionFusion} is suitable for filtering more difficult false positive and has better overall performance. Experimental results on the KITTI dataset demonstrate significant improvement in filtering false positive over the approach using only point cloud data. Furthermore, the proposed method can provide competitive results and has the fastest speed compared to the published state-of-the-art multi-modal methods in the KITTI benchmark.
\end{abstract}
	
%\vspace{-0.7em}	
%%%%%%%%%%%%%%%%%%%%%%%%%%%%%%%%%%%%%%%%%%%%%%%%%%%%%%%%%%%%%%%%%%%%%%%%%%%%%%%%
\section{INTRODUCTION}
\textbf{ Background. } Since 3D vehicle detection is a crucial part of perception in autonomous driving, a lot of research work has been invested in the industry and academia. 3D detection can be achieved through images, lidar point clouds or multi-modal data. In this study, we consider the fusion problem of lidar point clouds and RGB images. Point clouds provide very accurate depth information, but are accompanied by low resolution and texture information. On the other hand, images have ambiguous depth information but can provide fine-grained texture and color information. This provides an attractive research opportunity for how to design a model that can take full advantages of the two types of data.

\textbf{ Reality. } As a pioneering attempt of fusion methods, MV3D proposed a multi-view fusion mechanism to explore multi-modal collaboration methods. Based on this method, AVOD and ContFuse further proposed more diverse fusion pipelines to improve the performance of 3D detection. However, some recently published methods such as PointPillars\cite{lang2019pointpillars}, Second\cite{yan2018second}, PointRCNN\cite{shi2019pointrcnn}, PartA2\cite{shi2020points}, STD\cite{yang2019std} and 3DSSD\cite{yang20203dssd} significantly outperform these methods using only lidar point clouds. In fact, despite some recent fusion studies\cite{yoo20203d, liang2019multi, vora2020pointpainting, sindagi2019mvx, xie2020pi}, the top methods on the KITTI\cite{geiger2012we} leaderboard are still lidar only\cite{shi2020pv, liang2020rangercnn, he2020structure}. Does this mean that the 3D point cloud is sufficient for 3D detection? Or is there any way for RGB images to effectively supplement 3D detection?

\begin{figure}[tbp]
	\centering
	\includegraphics[width=0.99\linewidth]{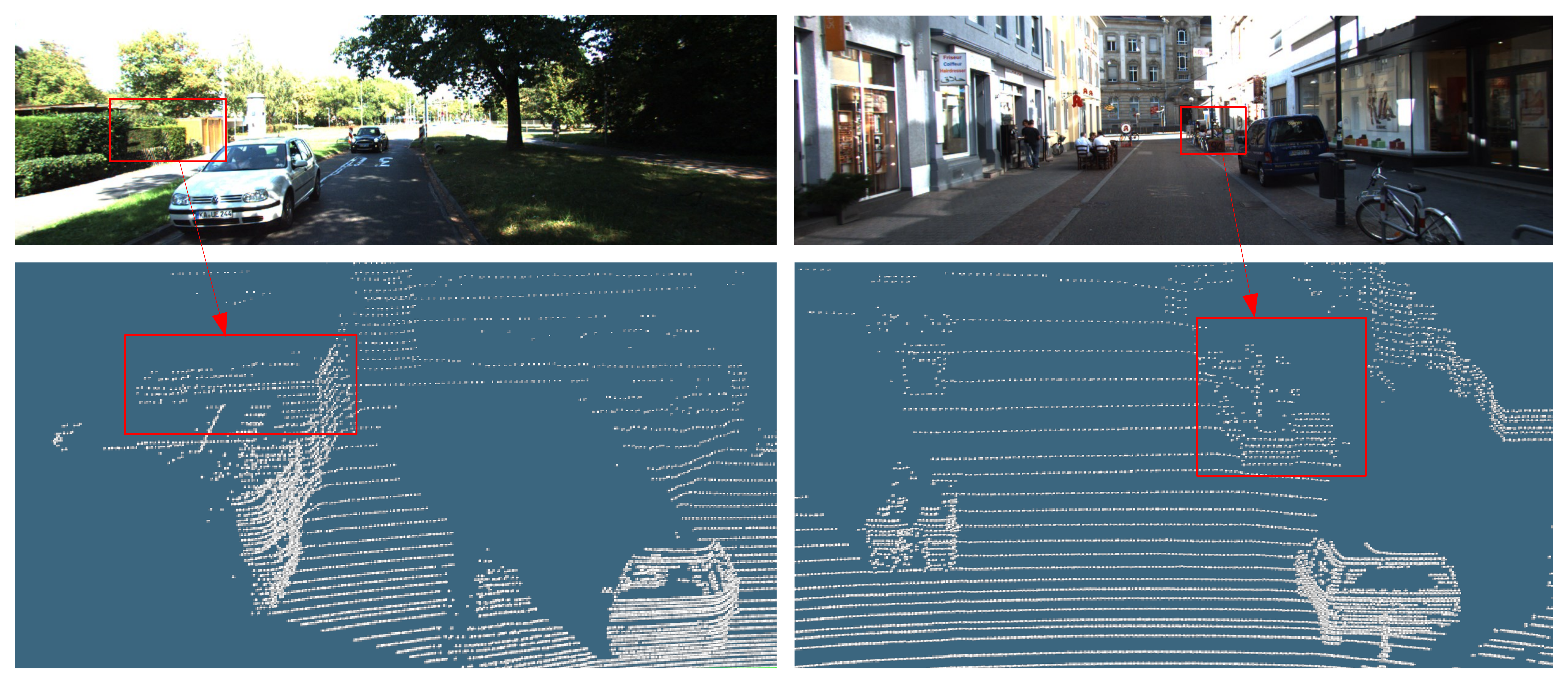}
	\caption{Sample scenes of LIDAR point clouds prone to false detection from KITTI dataset. The 3D structures of the hierarchical trees on the left and objects clusters on the right are similar to that of vehicles, so that they are easy to be misdetected as vehicles. The proposed MAFF-Net will adaptively add RGB image features to the point cloud network, making it easier to distinguish these ambiguous objects.}
%	\setlength{\belowcaptionskip}{-10cm}
%	\vspace{-0.4cm}
	\label{intro}
\end{figure}

\textbf{ Analysis. } The answer is clear by analyzing the characteristics of point cloud data and RGB image data. Consider the two examples in Fig. \ref{intro}, where the hierarchical trees in the left and the object clusters in the right are very similar to vehicles in the lidar modality. This makes it difficult to distinguish specific objects using only point clouds. But, the above objects are easy to distinguish in the image, which shows that adding images to the point cloud network can help effectively reduce 3D false positive(FP). However, in the currently published fusion methods, almost no attention has been paid to the specific effects of images for 3D detection, especially in eliminating FP. In this research, we focus on the false positive filtering effect of images for 3D vehicle detection: we hope that when the image is fused with the point cloud, FP can be effectively reduced without reducing true positive(TP).

\textbf{ Challenges. } However, how to effectively integrate 2D images into the 3D detection pipeline is still an open question. A naive way is to directly concatenate the raw RGB feature to the 3D point cloud feature, because the point-to-pixel correspondence can be established through projection. However, the image often has noise information such as occlusion and truncation. In these cases, the wrong image feature will be obtained after 3D points are projected onto the image. Therefore, simply using point-wise projected image features will degrade the performance of 3D detection. Moreover, we also notice that the detection speeds of current fusion methods are generally slow. The reasons for the slow speed are: (i) the fusion methods generally adopt  a two-stage way that uses ROI-pooling\cite{girshick2015fast} to fuse the image with the point cloud; (ii) the fusion methods usually use 2D detection or segmentation tasks to obtain high-level image features, and then fuse the features with the point cloud. The second way is equivalent to doing a 2D task first and then doing 3D task, which slows down the speed of the 3D detection.

\textbf{ Contributions. } In this paper, to address the above challenges, we design a multi-modal adaptive feature fusion network named MAFF-Net based on the successful PointPillars architecture. It is an end-to-end single-stage fusion 3D detection method, which only uses the raw RGB features and does not rely on the high-level features  of any 2D tasks. It is also able to eliminate the interference information of two  modalities through channel-wise attention module, so that the network can make full use of the advantages of image and point cloud features, to achieve the purpose that reducing FP and reserving TP at the same time.

Specifically, we have developed two fusion technologies: (i) \textit{PointAttentionFusion}(PAF): This is a relatively fast and simple fusion method where the corresponding projection image features and 3D point features are concatenated to obtain preliminary mixed features, which are used as the input of point-wise channel attention module to learn the channel-wise importance of point cloud and image features. The learned attention features and the original modal features are then concatenated to obtain the final point-wise fusion feature, which is then jointly processed by the PointPillars pipelines. (ii) \textit{DenseAttentionFusion}(DAF): In this technology, point cloud and image features are divided into three types of pillar-wise features: point cloud feature, image feature, and point cloud image feature that is obtained by appending RGB feature to the point cloud feature. The above three features are concatenated and sent to the pillar-wise channel attention module to generate attention feature. Finally, the attention feature is concatenated with the previous three modalities features to form the pillar-wise fusion feature, which is further used by the subsequent 2D CNN network. Compared with the \textit{DenseAttentionFusion}, \textit{PointAttentionFusion} is more concise, so it is faster and more suitable for eliminating some simple false detections. \textit{DenseAttentionFusion} is a complex and dense fusion method. Its performance is better than \textit{PointAttentionFusion} and it is more effective for some difficult false detections.

The main contributions of this study are summarized as follows:
\begin{itemize}
	\item This paper proposes an end-to-end trainable single-stage multi-modal fusion method for reducing 3D false positive.

	\item This paper proposes two different fusion techniques based on channel attention mechanism for different usage scenarios.
	
	\item Experiments on the KITTI dataset demonstrate that the proposed method can effectively reduce false positive while maintaining true positive. It achieves competitive results compared with the published state-of-the-art fusion methods with a fast detection speed of 32 Hz.
	
	\item Extensive analysis and visualization are conducted to understand the design principles of the proposed method and the usefulness of the attention mechanism for multi-modal fusion.

\end{itemize}

\section{RELATED WORK}
\subsection{3D Vehicle Detection based on Point Cloud}

3D vehicle detection using only point clouds can be roughly divided into 3 types: Bird’s Eye View(BEV)-based, Voxel-based, and Point-based. BEV-based methods\cite{simony2018complex, redmon2017yolo9000} first use rules\cite{yang2018pixor} to convert the point cloud into the BEV form, and then uses the 2D CNN network for learning and prediction\cite{zeng2018rt3d}. Voxel-based methods divides the point cloud space into regularly arranged voxel blocks, use rules\cite{wang2015voting, engelcke2017vote3deep, li20173d} or neural networks\cite{zhou2018voxelnet, yun2019focal} to encode the voxels, and then adopt 3D or 2D convolution\cite{yan2018second, yi2020segvoxelnet, shi2020points, liu2020tanet, ye2020hvnet} for 3D detection. PointPillars\cite{lang2019pointpillars} further simplified voxels into pillars, realizing a real-time one-stage 3D detection method. The voxel-based methods are conducive to accurate 3D proposal generation but are also limited by the receptive field of 2D/3D convolution. The point-based method mainly uses PointNet\cite{qi2017pointnet, qi2017pointnet++} technology to encode 3D points, and then uses more strategies\cite{shi2019pointrcnn, yang2019std, yang20203dssd} to process 3D points to achieve the purpose of accurately predicting 3D vehicles. Most of these point-based methods are based on the PointNet series, which makes the receptive fields of point cloud feature learning more flexible. Recently, there has been a trend of fusing point-based and voxel-based methods\cite{he2020structure, shi2020pv} to utilize the best of two worlds, thereby improving 3D detection performance. 

\subsection{3D Vehicle Detection based on Multi-modal Fusion}

\begin{figure*}[htbp]
	\centering
	\includegraphics[width=0.99\linewidth]{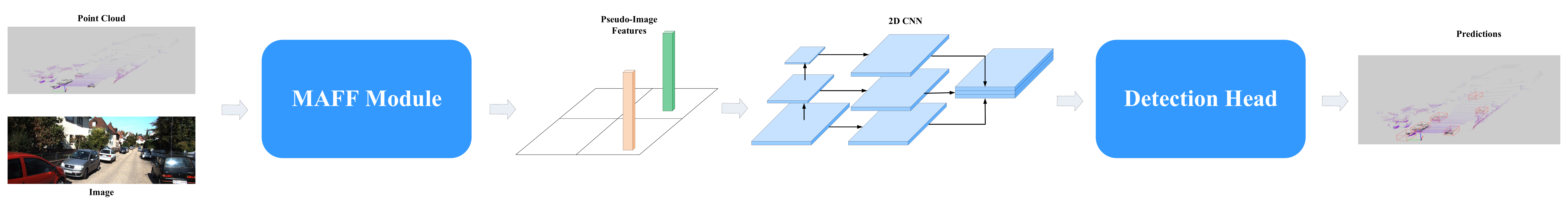}
	\caption{Overview of the proposed MAFF-Net. The net is an end-to-end point cloud and image fusion method. It uses point cloud and image as input, outputs a set of pillar-wise fusion features through the MAFF module, and then scatter the fusion features back to a 2D pseudo-image for a 2D convolutional neural network according to the spatial relationship of the pillars. The features from the 2D backbone are used by the detection head to predict 3D bounding boxes for vehicles. }
%	\vspace{-0.4cm}  
	\label{whole_structure}
\end{figure*}

In order to take advantages of the camera and lidar sensors, various fusion methods have also been proposed. MV3D\cite{chen2017multi} is a pioneering work in the fusion method. It combines the lidar BEV representation, front view and image together, and proposes a two-stage network. AVOD\cite{ku2018joint} fuses the features of the BEV and the image in the middle layer of the convolution to predict 3D objects. ContFuse\cite{liang2018deep} uses continuous convolution to fuse images and lidar features on different resolutions. MMF adds ground estimation and depth estimation to the fusion framework, and learns better fusion feature representations while jointly learning multi-tasks. \cite{qi2018frustum, wang2019frustum, shin2019roarnet} first use camera images to generate proposals and then exploit some methods to process the lidar points in these regions to generate 3D objects. \cite{sindagi2019mvx, dou2019seg} use 2D tasks (detection or segmentation) to obtain the feature representations of the image, then fuse these feature representations with the point cloud, and finally use the architecture of VoxelNet to process the fused features.

Although various multi-modal fusion methods have been proposed, they have seldom paid attention to the specific gains brought by the image for 3D detection and the detection speeds are often very slow. In this study, we focus on using images to eliminate the FP of 3D vehicle detection and hope to obtain a fast multi-modal fusion method.

%%%%%%%%%%%%%%%%%%%%%%%%%%%%%%%%%%%%%%%%%%%%%%%%%%%%%%%%%%%%%%%%%%%%%%%%%%%%%%%%
\section{Proposed Method}
In this section, we introduce the proposed single-stage point cloud and image feature fusion 3D vehicle detector MAFF-Net in detail.

\subsection{Raw RGB Input}

Compared with lidar point clouds, images have rich color and texture information, which is very effective for overcoming the false detection of point clouds, especially the background false detection. In this paper, we only use the raw RGB features of the image as the input of MAFF-Net. MAFF-Net does not rely on any 2D annotation information and the high-level features of any 2D tasks. 

\subsection{PointPillars}

The PointPillars structure is used as the base 3D detection network for two main reasons: (i) it achieves a good balance between speed and performance: with excellent performance, it has a very fast detection speed. A lot of work in academia and industry is also based on this algorithm\cite{liu2020tanet, ye2020hvnet, kuang2020voxel, caesar2020nuscenes, zhou2020end, wang2020pillar}. (ii) It provides a natural and effective interface for fusing image features at different granularities in 3D space such as points and pillars. The network used in this study is described in \cite{lang2019pointpillars}. For completeness, this section briefly reviews PointPillars. This algorithm consists of three modules: (i) a Pillar Feature Net(PFN); (ii) 2D Convolutional Neural Networks; (iii) a Detection Head.

PFN is a feature learning network designed to encode raw point clouds for individual pillars. All non-empty pillars are encoded by PFN and share the same network parameters. The encoded features are scattered back to the original pillar positions to construct a pseudo-image. The pseudo-image features are forwarded through a series of 2D convolutional blocks to extract high-level features. The features are then used by a detection head to generate the targets.

\subsection{Multimodal Adaptive Feature Fusion(MAFF)}
In this study, two concise technologies that can fuse raw RGB data with the point cloud data are proposed to filter false positive in 3D detection. 

\begin{figure*}[htbp]
	\centering
	\includegraphics[width=0.83\linewidth,height=0.27\linewidth]{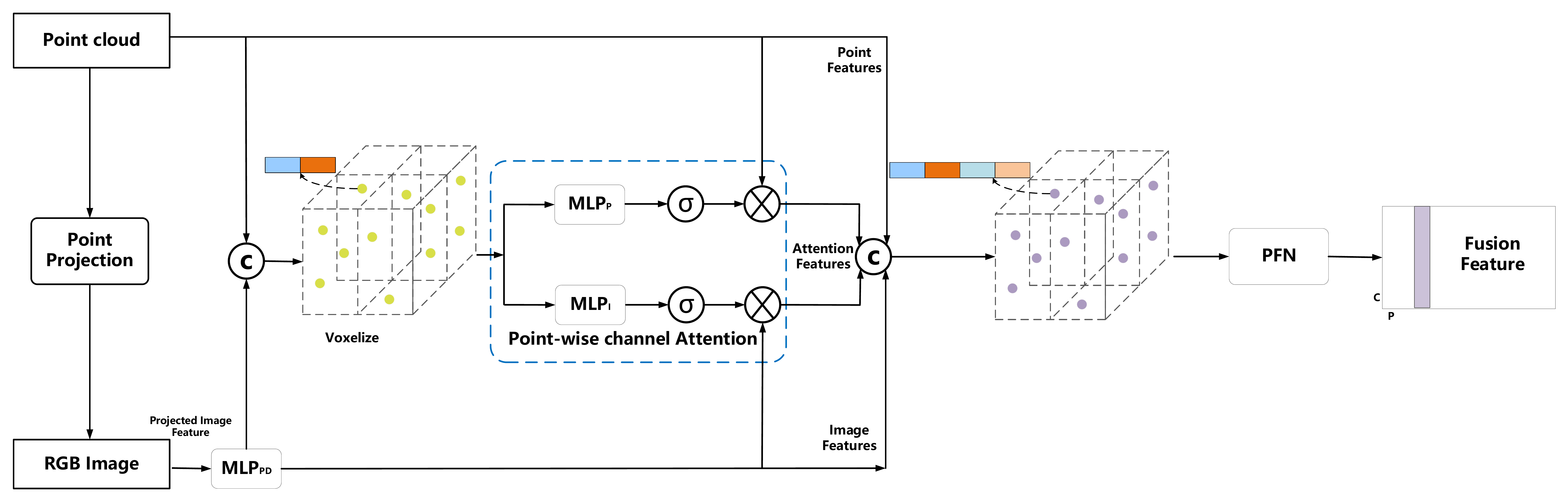}
	\caption{Architecture of the proposed MAFF module with \textit{PointAttentionFusion} technology. Different colors represent the features of different modalities}
	\label{point_fusion}
%	\vspace{-0.4cm}
\end{figure*}

\textbf{ \textit{PointAttentionFusion(PAF):}} This is an early and simple fusion approach in which each 3D point is aggregated through an image feature and two attention features to achieve the adaptive fusion features. Fig. \ref{point_fusion} presents the procedure of this technology.

The method first uses the calibration matrix to project each 3D point onto the image, and the projected image features can be obtained according to the corresponding projection location index. Note that these features are the raw RGB features.  In order to make the projected image features and point cloud features compatible, a simple fully connected network called $\mathrm{MLP_{PD}}$ is applied to map the image features to appropriate dimensions. The $\mathrm{MLP_{PD}}$ is composed of a set of blocks and each block consists of a linear layer, a BN layer, and a ReLU layer. Next, the point cloud features and the mapped image features are concatenated channel-wise to obtain point-wise extended features. However, since the image has many noise factors, such as occlusion, truncation, etc., the extended features will introduce interference information.

To address this issue, we adopt the point-wise channel attention module, which uses expanded features to adaptively estimate the importance of each type of features in a channel-wise manner. First, feed the extended features into a fully connected layer that includes a linear layer, a ReLU layer, and a linear layer. Then output the feature weights through a sigmoid, and finally multiply the weights with the corresponding features in a channel-wise manner to get the attention features. The above process is used to process the point cloud features and image features respectively. The specific forms of channel attention are as follows:
\begin{equation}
\begin{aligned}
\mathbf{F_{a.P}} &= \mathbf{F_P} \otimes \underbrace{\sigma(\mathrm{MLP_P}(\mathbf{F_E}))}_{\text{Point cloud attention}} \\
\mathbf{F_{a.I}} &= \mathbf{F_I} \otimes \underbrace{\sigma(\mathrm{MLP_I}(\mathbf{F_E}))}_{\text{Image attention}}  \\ 
\end{aligned}
\end{equation}
where $\mathbf{F_P}$ and $\mathbf{F_I}$ represent point cloud and image features, respectively, $\mathbf{F_{a.P}}$ and $\mathbf{F_{a.I}}$ are the corresponding point-wise attention features, $\mathbf{F_E}$ is the extended point image features, $\sigma$ is the sigmoid activation function and $\otimes$ is the element-wise product operator. After obtaining the attention features, concatenate $\mathbf{F_P}$, $\mathbf{F_I}$, $\mathbf{F_{a.P}}$ and $\mathbf{F_{a.I}}$ channel-wise to get the point-wise fusion feature. Then divide the 3D point cloud space into pillars, followed by grouping the points to pillars. Finally, the simplified version of PointNet is adopted to generate the pillar-wise fusion features.

The advantage of this method is that since the fusion method is simple and the added networks are all shallow networks, the approach can achieve fast detection speed. Moreover, the approach can learn to summarize useful information from both modalities through PFN layer because of the early stage fusion strategy.

\begin{figure*}[htb]
	\centering
	\includegraphics[width=0.85\linewidth,height=0.3\linewidth]{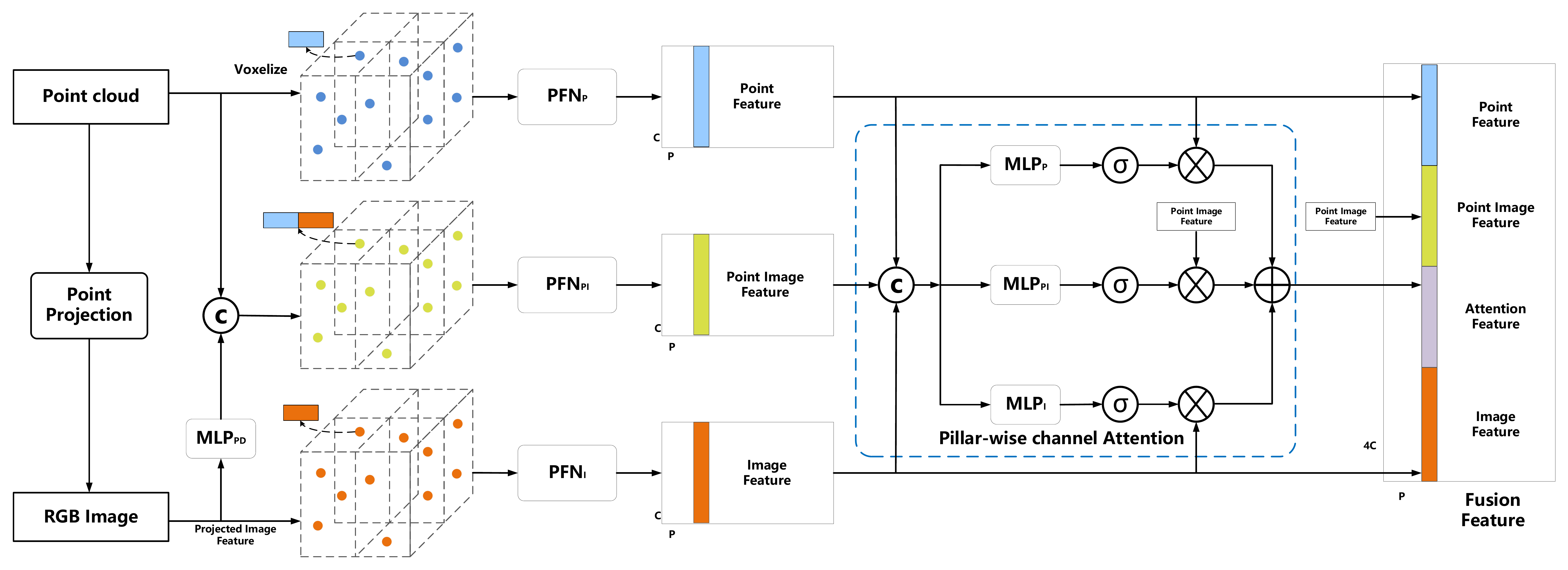}
	\caption{Architecture of the proposed MAFF module with \textit{DenseAttentionFusion} technology. Different colors represent the features of different modalities}
%	\vspace{-0.4cm}
	\label{dense_fusion}
\end{figure*}

\textbf{ \textit{DenseAttentionFusion(DAF):}} In contrast to \textit{PointAttentionFusion} that fuses features in a concise manner, \textit{DenseAttentionFusion} employs a relatively complex fusion strategy, where features are divided into three forms and fused together. As shown in Fig. \ref{dense_fusion}, the three features are the point cloud features, the image features, and the extended point image features $\mathbf{F_E}$ described in \textit{PointAttentionFusion}.

In \textit{PointAttentionFusion}, only extended point image features $\mathbf{F_E}$ is used for subsequent work. However, after PFN learning, the feature is equivalent to the global features of the point cloud and the image. This feature will lose some characteristics of the original two modalities. Thus, a naive approach would be to put the original point cloud features and image features back into the global features. However, due to the noise information of the image features and the redundancy of the three features, blindly fusing the three features will introduce lots of interference information, resulting in the degradation of detection performance. In the following, we describe a novel dense attention fusion method that effectively combines the three types of features.

First, after projecting 3D points onto the image, the corresponding RGB features can be obtained. The RGB features have two uses, one is to expand the point cloud features in the same way as \textit{PointAttentionFusion} to obtain the point cloud image features, and the other can regard this point-wise image feature as another feature form of 3D points. In this way, we can obtain point cloud features, point cloud image mixed features, and projected image features. Because the above three features are all in point-wise form, they can use the $x,y$ and $z$ coordinates of the 3D point to perform pillar operations, respectively. Here, we can get the pillar representations of three different features, and then three PFNs with the same structure are adopted to encode the pillars to generate three type of pillar-wise features.

Next, we use the pillar-wise channel attention module to adaptively estimate the importance of each pillar feature. The above three features are concatenated together as the input of this module. The attention features of the three features are estimated by the following attention maps:
\begin{equation}
\begin{aligned}
\mathbf{F_{a.P}} &= \mathbf{F_P} \otimes \underbrace{\sigma(\mathrm{MLP_P}(\mathbf{F_C}))}_{\text{Point cloud attention}} \\
\mathbf{F_{a.PI}} &= \mathbf{F_{PI}} \otimes \underbrace{\sigma(\mathrm{MLP_{PI}}(\mathbf{F_C}))}_{\text{Point Image attention}} \\
\mathbf{F_{a.I}} &= \mathbf{F_I} \otimes \underbrace{\sigma(\mathrm{MLP_I}(\mathbf{F_C}))}_{\text{Image attention}}  \\
\mathbf{F_{A}} &= \mathbf{F_{a.P}} + \mathbf{F_{a.PI}} + \mathbf{F_{a.I}}\\ 
\end{aligned}
\end{equation}
where $\mathbf{F_P}$, $\mathbf{F_{PI}}$ and $\mathbf{F_I}$ represent the pillar-wise point cloud, point image and image features, respectively,  $\mathbf{F_C}$ is the concatenated feature of the above three features, $\mathbf{F_{a.P}}$, $\mathbf{F_{a.PI}}$ and $\mathbf{F_{a.I}}$ are the corresponding pillar-wise attention features, $\mathbf{F_{A}}$ is the pillar-wise attention features that need to be obtained, $\sigma$ is the sigmoid activation function and $\otimes$ represents the element-wise product operator. After obtaining the attention features, concatenate $\mathbf{F_P}$, $\mathbf{F_{PI}}$, $\mathbf{F_I}$, and $\mathbf{F_{A}}$ channel-wise to get the pillar-wise fusion features.

Although \textit{DenseAttentionFusion} is a relatively complex fusion strategy, it has the following advantages. First, it fully retains the original characteristics of the three features of point cloud, image, and point cloud image, while minimizing the noise impact of each feature. This makes its detection performance better than \textit{PointAttentionFusion}. Second, the fusion is based on Pillars features, which can reduce the dependence on the availability of high-resolution 3D points.

\subsection{Training Details}
\textbf{ Network Architecture: } For the fairness of comparison, we keep most of the settings of PointPillars as described in \cite{lang2019pointpillars} except for some newly added module structures. According to the type of fusion, the input and output dimensions of the PFN layers and the MLP layers are different. Note that, except for the input dimensions, the structure of 2D convolutional network in the proposed MAFF-Net is the same as PointPillars.

For \textit{PointAttentionFusion}, the configuration of image dimension prediction $\mathrm{MLP_{PD}}$ is (3,96,16). The $\mathrm{MLP_P}$ and $\mathrm{MLP_I}$ in the attention module have (9+16,25,9) and (9+16,25,16) configurations respectively, where 9 is the dimension of the point cloud feature and 16 is the dimension of the image feature. Next, the configuration of PFN layer is (9+16+9+16,64), which leads to the input dimension of the subsequent 2D convolutional network is 64.

For \textit{DenseAttentionFusion}, the configuration of $\mathrm{MLP_{PD}}$ is the same as the $\mathrm{MLP_{PD}}$ in \textit{PointAttentionFusion}. $\mathrm{PFN_P}$, $\mathrm{PFN_{PI}}$ and $\mathrm{PFN_I}$ have configurations of (16, 64), (9+16, 64) and (3, 64) respectively. The configuration of the three MLPs in the attention module are all (64*3, 64*3, 64), but their weights are not shared. The final fusion feature is combined by four types of features, so its dimension is 64*4, which leads to the input dimension of the 2D convolutional network is 64*4.

\textbf{ Loss: } We use the same loss functions described in PointPillars. The loss function is divided into 3 types, namely the localization regression loss $\mathcal L_{loc}$, the classification loss $\mathcal L_{cls}$, and the orientation loss $\mathcal L_{dir}$. $\mathcal L_{loc}$ uses SmoothL1 function to define the loss for  $(x,y,z,w,l,h,\theta)$, $\mathcal L_{cls}$ uses focal loss, and $\mathcal L_{dir}$ uses a softmax classification loss. The overall loss function can be defined as:
\begin{equation}
\mathcal L_{total}  =\frac{1}{{N_{pos} }}\left( {\beta _{loc} \mathcal L_{loc} {\rm{ + }} \beta _{cls} \mathcal L_{cls} {\rm{ + }} \beta _{dir} \mathcal L_{dir} } \right)
\end{equation}
where $\beta _{loc}=2.0,\ \beta _{cls}=1.0,\ \beta _{dir}=0.2$, and $N_{pos}$ is the number of positive anchors.

\textbf{ Data Augmentation: } Our two fusion methods both project the 3D points to the image in a point-wise manner, so all the data augmentation methods of PointPillars can be used, which is similar to PointPainting\cite{vora2020pointpainting}. Data augmentation adopts sample objects from database, augment ground truths independently, and perform global augmentations for the whole point cloud and all boxes.

\section{EXPERIMENTS}
\subsection{Dataset and Metric}

\textbf{ Dataset and Implementation: } This study uses the KITTI benchmark dataset \cite{geiger2012we} to evaluate the proposed fusion method, which includes 7481 training samples and 7518 testing samples. There are three levels of difficulty: easy, moderate and hard, which are assessed based on the object height in image, occlusion and truncation. The rank of leaderboard is based on the moderate result. We follow the general approach proposed by \cite{chen2017multi} to split the training samples into a training set of 3712 samples and a validation set of 3769 samples. The proposed MAFF-Net is compared with previously published method on the car category. Inference of the entire network is carried out on Tesla V100 GPU. The code of PointPillars used in this study comes from (httP://github.com/nutonomy/second.pytorch).

\textbf{ Metric: } The metric of KITTI is defined by average precision(AP) on 40 recall positions of the Precision/Recall curve \cite{simonelli2019disentangling} with IoU=0.7. This metric is a good indicator of the overall performance, but it cannot effectively reflect the details of the performance, such as the FP that we are concerned about. To show the performance of filtering FP, a simple way can be used: for each recall position, the number of TP is the same. According to the formula of precision $precision = \frac{TP}{TP+FP}$, when TP remains unchanged, the less FP, the higher precision. Thus, the precision at a single recall position can reflect the performance of the algorithm processing FP. In this study, we use precision of five recall positions ($0.725,0.75,0.775,0.8$) to verify the proposed method performance for filtering FP. The reason for choosing these positions is that the scores of detection at these positions are generally at a medium level and it is difficult to filter vehicles only using the score threshold, which will lead to many FP. 

\begin{table*}[h]
	%	\centering
	%	\large
%	\vspace{0.5cm}
	\caption{3D detection performance(\%) comparison on the KITTI \textit{validation} set (Car). Top-1 method is highlighted in BOLD.}
%	\vspace{-0.3cm}
	\resizebox{\textwidth}{!}{%
		\begin{tabular}{c|cccc|cccc|cccc|cccc}
			\hline
			\multirow{2}{*}{Method} & \multicolumn{4}{c|}{Recall=0.725} & \multicolumn{4}{c|}{Recall=0.75} & \multicolumn{4}{c|}{Recall=0.775} & \multicolumn{4}{c}{Recall=0.8}                                   \\
			& easy & moderate & hard & mAP & easy & moderate & hard & mAP & easy           & moderate       & hard           & mAP            & easy           & moderate       & hard           & mAP            \\ \hline
			PointPillars & 95.17 & 86.21 & 81.75 & 87.71 & 94.29 & 83.68 & 77.41 & 85.13 & 93.49          & 79.99          & 71.39          & 81.62          & 92.41          & 73.86          & 66.03          & 77.44          \\
			MAFF-Net(PAF)  & \textbf{96.48} & 87.56 & 82.83 & 88.96 & \textbf{96.16} & 85.25 & 78.91 & 86.78  & \textbf{95.34} & 81.58          & 73.31          & 83.41 & \textbf{94.44} & 75.69          & 67.99          & 79.37          \\
			MAFF-Net(DAF)  & 95.64 & \textbf{88.35} & \textbf{83.53} & \textbf{89.17} & 95.03 & \textbf{86.08} & \textbf{80.11} & \textbf{87.07}  & 94.23 & \textbf{83.05} & \textbf{73.56} & \textbf{83.61}          & 93.49          & \textbf{78.08} & \textbf{68.82} & \textbf{80.13} \\
			\textit{Improvement}   & \textit{+1.31} & \textit{+2.13} & \textit{+1.78} & \textit{+1.46} & \textit{+1.87} & \textit{+2.40} & \textit{+2.69} & \textit{+1.94} & \textit{+1.86} & \textit{+3.06} & \textit{+2.17} & \textit{+1.99} & \textit{+2.03} & \textit{+4.22} & \textit{+2.79} & \textit{+2.69} \\ \hline
		\end{tabular}%
	}
%	\vspace{-0.3cm}
	\label{filter}
\end{table*}

\subsection{Evaluation on KITTI Validation Set}
\label{ex_eval}

\textbf{ Evaluation on filtering FP. }Table \ref{filter} shows the comparison of the 3D detection performance of PointPillars and MAFF-Net at different recall positions. The performance of each position has been significantly improved. As recall increases (in this case, the detection score decreases and FP is easier to increase), the improvement of mAP gradually increases. For the moderate category that is concerned on the KITTI benchmark, when the recall is 0.8, the proposed fusion method improves the performance by 4.22\%. For a single recall position, only reducing FP can improve performance because of the same number of TP. Therefore, the performance improvement on each single recall position shows that MAFF-Net is effective for filtering FP.

\begin{table}[h]
	\normalsize
	\caption{The number of TP and FP for 3D detection under different score thresholds (Car). BG means background. Top-1 method is highlighted in BOLD.}
%	\vspace{-0.3cm}
	\resizebox{0.5\textwidth}{!}{%
		\begin{tabular}{ccccccc}
			\toprule
			\multirow{2}{*}{Method} & \multicolumn{3}{c}{Score Threshold = 0.4} & \multicolumn{3}{c}{Score Threshold = 0.1} \\
			& TP     & FP             & FP(BG)  & TP    & FP              & FP(BG) \\ \midrule
			PointPillars & 8606   & 4428           & 2346            & 8783  & 26237           & 22403          \\
			MAFF-Net(PAF)    & \textbf{8636}   & 4080           & 2018            & 8802  & 24446           & 20478          \\
			MAFF-Net(DAF)    & 8627   & \textbf{3933}  & \textbf{1906}   & \textbf{8811}  & \textbf{23330}  & \textbf{19585} \\
			\textit{Improvement number/rate}    & -      & -495          & -440            & -     & -2907         & -2818          \\ \bottomrule
		\end{tabular}%
	}
%	\vspace{-0.5cm}
	\label{fp_number}
\end{table}

In order to reflect the performance of filtering FP more intuitively, Table \ref{fp_number} lists the number of TP and FP for the 3D vehicle detection under different score thresholds. When the score are 0.4 and 0.1, the FP are reduced by 11.18\% and 11.08\%, respectively, and the FP caused by the background are reduced by 18.75\% and 12.58\%, respectively. It shows that the proposed two fusion methods can effectively reduce FP while slightly improving TP, especially for reducing FP caused by background.

%\textbf{ Comparison with state-of-the-art. } Table \ref{compare_validation} shows the AP scores of the proposed two fusion methods and other state-of-the-art methods on the KITTI validation set using 3D and bird's eye view(BEV) evaluation. Among the most published 3D detection methods based on multimodal fusion, the method proposed in this study has excellent detection performance, especially in the BEV evaluation. At the same time, the running speeds of the two proposed fusion methods are also very fast in the fusion methods. \textit{PointAttentionFusion} and \textit{DenseAttentionFusion} reach 32 Hz and 24 Hz, respectively.
%
%It can also be obsered from Tables \ref{filter} and \ref{compare_validation} that the overall performance of \textit{PointAttentionFusion} is slightly lower than that of \textit{DenseAttentionFusion}, because \textit{DenseAttentionFusion} fully integrates the features of three forms. It is worth pointing out that \textit{PointAttentionFusion} has a faster running speed because of the simple fusion form. And, in the easy category, \textit{PointAttentionFusion} is better than \textit{DenseAttentionFusion} in terms of the ability to filter FP according to Table \ref{filter}. Therefore, the two fusion methods proposed in this study are suitable for different situations, and researchers can choose different fusion methods according to actual needs.

\begin{table}[h]
%		\vspace{-0.2cm}
	\caption{Comparison of 3D and BEV AP with/without attention module on the KITTI \textit{validation} set (40 recall positions).}
%	\vspace{-0.3cm}
	\resizebox{0.5\textwidth}{!}{%
		\begin{tabular}{c|ccc|ccc}
			\hline
			\multirow{2}{*}{Method} & \multicolumn{3}{c|}{3D(Car)}  & \multicolumn{3}{c}{BEV(Car)}  \\
			& easy  & moderate & hard  & easy  & moderate & hard  \\ \hline
			PointPillars(baseline)           & 87.79 & 78.44    & 74.06 & 92.55 & 88.32    & 86.52 \\
			MAFF-Net(PointFusion)             & 89.02 & 77.76    & 74.51 & 93.55 & 87.52    & 84.78 \\
			MAFF-Net(PAF)    & 88.86 & 79.30    & 74.71 & 93.10 & 89.25    & 86.67 \\
			MAFF-Net(DenseFusion)             & 88.05 & 77.20    & 73.73 & 93.10 & 87.21    & 84.29 \\
			MAFF-Net(DAF)    & 88.88 & 79.37    & 74.68 & 93.23 & 89.31    & 86.61 \\ \hline
		\end{tabular}%
	}
	\label{attention_table}
%		\vspace{-0.4cm}
\end{table}

\begin{figure}[h]
	\centering
	\includegraphics[width=0.99\linewidth]{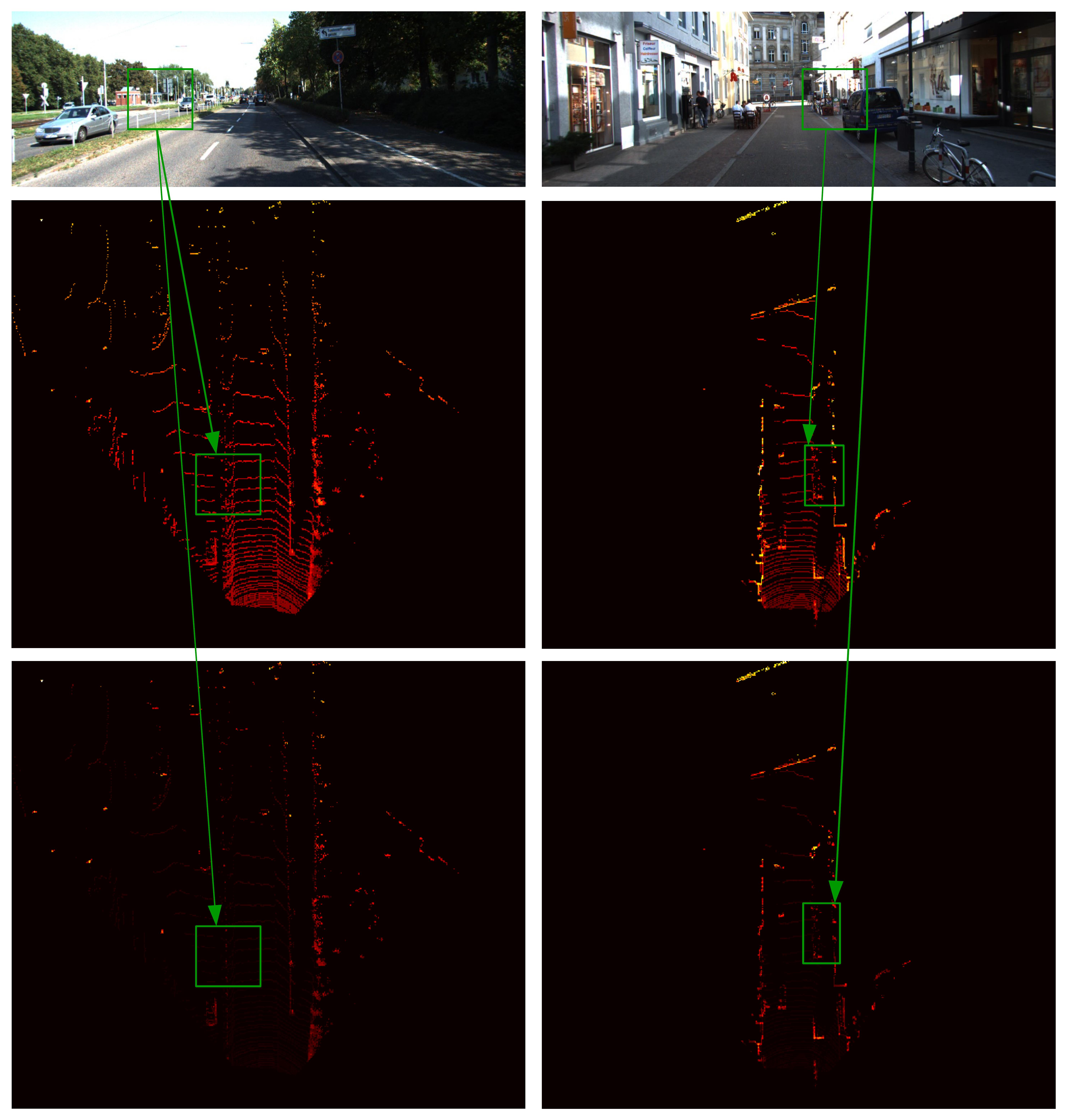}
	\caption{Visualization of features before/after attention. First line(top): image, Second line: point cloud feature before attention in \textit{DenseAttentionFusion}, Third line: point cloud feature after attention in \textit{DenseAttentionFusion}. Green rectangular draws the area that is easy to be false detected. The attention module can effectively suppress the background area and enhance the shape of the vehicle.}
%	\vspace{-0.5cm}
	\label{attention}
\end{figure}

\textbf{ Analysis of the attention mechanisms. } We visualize the point cloud features in \textit{DenseAttentionFusion} before ($\mathbf{F_{P}}$) and after ($\mathbf{F_{a.P}}$) using attention, as shown in Figure \ref{attention}. After using the attention module, the background area is fully suppressed, so that the geometric shape of the vehicles can be highlighted. Moreover, for some objects that are similar to the vehicle in 3D structure, the attention module can eliminate some of their shapes, so that their 3D structures are no longer similar to the vehicle.

In order to further show the role of the attention module, Table \ref{attention_table} shows the performance comparison of the KITTI validation set with and without the attention module. When the attention module is not used, compared to PointPillars, the two fusion methods have improved performance in the easy category, but the performance in the moderate and hard categories decreases. In the moderate and hard categories, interference information is very serious. In these cases, when the point cloud is projected to the image, it will get the wrong image features, which will lead to the degradation of detection performance. After adding the attention module, the MAFF-Net is able to adapt to the weight of each modal feature, thereby improving the detection performance.

\begin{figure*}[htb]
	\centering
	\includegraphics[width=0.99\linewidth]{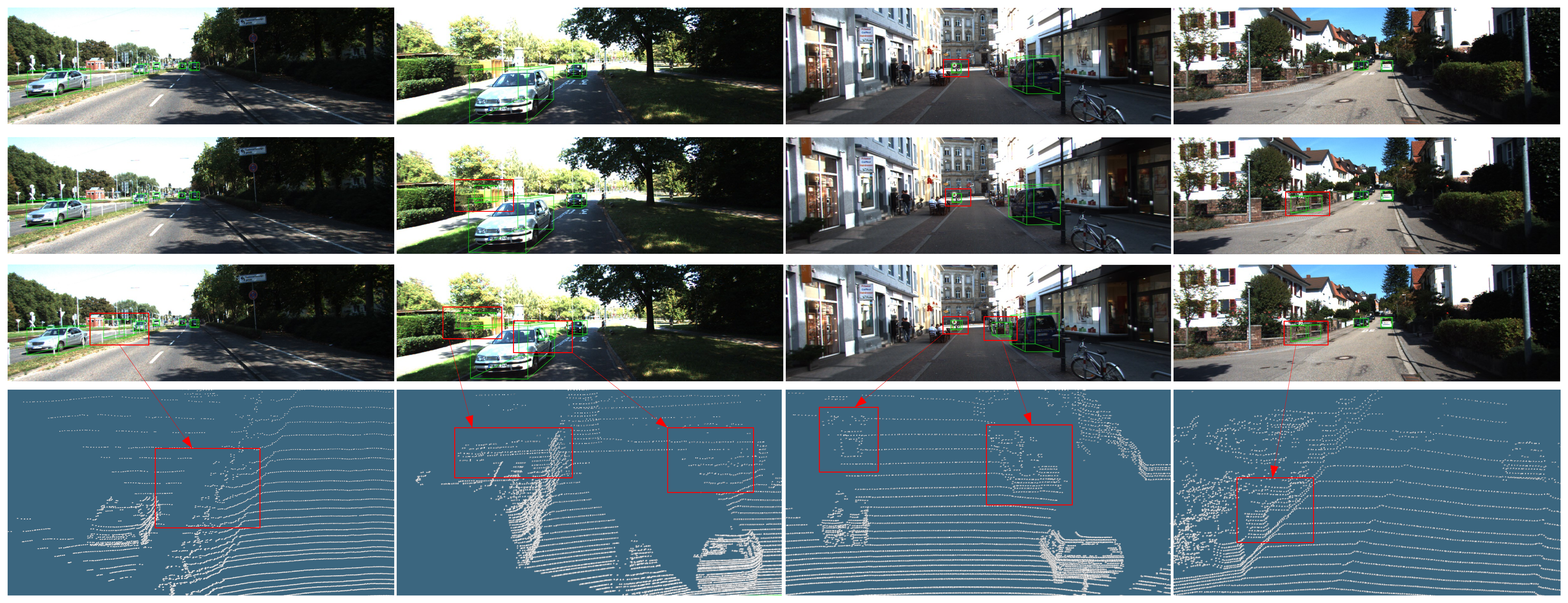}
	\caption{Qualitative analysis of results from KITTI validation dataset by projecting 3D bounding boxes  into image for clearer visualization. Fist line(top): MAFF-Net with \textit{DenseAttentionFusion}, Second Line: MAFF-Net with \textit{PointAttentionFusion}, Third line: PointPillars(baseline), Last Line(bottom): Spatial distribution of 3D point cloud. Green 3D boxes indicate prediction results. Red rectangles highlight false positive.}
%	\vspace{-0.5cm}
	\label{prediction}
\end{figure*}

\textbf{ The visualization of Detection Results. } Figure \ref{prediction} provides a qualitative analysis of some 3D detection results. It can be observed from Figure \ref{prediction} that the 3D structures of many false detection objects are similar to that of vehicles, such as the hierarchical tree clusters on the left side of the second column image, the object clusters on the right side of the third column image, and wall and fence on the left side of the fourth column image. These conditions will make it difficult for the lidar point cloud to distinguish the objects. However, MAFF-Net can effectively reduce the above false detection objects by using image information, which demonstrates the effectiveness of the proposed algorithm.

\subsection{Evaluation on KITTI Testing Set} We evaluate the proposed MAFF-Net with \textit{DenseAttentionFusion} on the KITTI testing set by submitting the test results to the official server. The results are summarized in Table \ref{testing}. It can be observed that MAFF-Net with \textit{DenseAttentionFusion} can have competitive results compared with other state-of-the-art multimodal fusion algorithms. The proposed method has the fastest speed, achieves top rank in one category and 2nd rank in three categories within the fusion methods. It is worth pointing out that the results of the test set we submitted(PointPillars(baseline)) using the code given in PointPillars\cite{lang2019pointpillars} are not ideal, for example, it is lower than the original PointPillars 1.1\% in the 3D moderate category. However, even if only the low-performance PointPillars are used, the detection performance of MAFF-Net is still better than the original PointPillars in most categories. In 3D moderate, MAFF-Net is 1.9\% higher than PointPillars (baseline) and 0.7\% higher than PointPillars. Therefore, if a more powerful point cloud backbone network can be used, MAFF-Net will achieve better performance.
\begin{table}[h]
%	\vspace{-0.3cm}
\caption{Comparison of 3D and BEV AP with other multimodal published methods on the KITTI \textit{testing} set. Top-2 methods are highlighted in BOLD.}
%\vspace{-0.3cm}
	\resizebox{0.5\textwidth}{!}{%
		\begin{tabular}{c|cccccc|c}
			\hline
			\multirow{2}{*}{Method} & \multicolumn{3}{c}{3D(Car)}                                                & \multicolumn{3}{c|}{BEV(Car)}                         & Speed \\
			& easy           & moderate       & hard                                & easy           & moderate       & hard           & (Hz)  \\ \hline
			MV3D\cite{chen2017multi}                    & 74.97          & 63.63          & \multicolumn{1}{c|}{54.00}          & 86.62          & 78.93          & 69.80          & 2.80  \\
			AVOD-FPN\cite{ku2018joint}               & 83.07          & 71.76          & \multicolumn{1}{c|}{65.73}          & 90.99          & 84.82          & 79.62          & 10.00 \\
			ContFuse\cite{liang2018deep}                & 83.68          & 68.78          & \multicolumn{1}{c|}{61.67}          & \textbf{94.07} & 85.35          & 75.88          & 16.70 \\
			F-ConvNet\cite{wang2019frustum}              & \textbf{87.36} & \textbf{76.39} & \multicolumn{1}{c|}{66.69}          & \textbf{91.51} & 85.84          & 76.11          & 2.1   \\
			F-pointnet\cite{qi2018frustum}              & 82.19          & 69.79          & \multicolumn{1}{c|}{60.59}          & 91.17          & 84.67          & 74.77          & 5.90  \\
			PI-RCNN\cite{xie2020pi}                 & 84.37          & 74.82          & \multicolumn{1}{c|}{\textbf{70.03}} & 91.44          & 85.81          & \textbf{81.00} & 10.00 \\
			PointPillars\cite{lang2019pointpillars}            & 82.58          & 74.31          & \multicolumn{1}{c|}{\textbf{68.99}} & 90.07          & 86.56          & \textbf{82.81} & 62.00 \\
			PointPillars(baseline)  & 83.11          & 73.12          & \multicolumn{1}{c|}{67.73}          & 90.06          & \textbf{86.64} & 79.19          & 62.00 \\
			MAFF-Net(DAF)           & \textbf{85.52} & \textbf{75.04} & \multicolumn{1}{c|}{67.61}          & 90.79          & \textbf{87.34} & 77.66          & 24.00 \\ \hline
		\end{tabular}%
	}
%\vspace{-0.3cm}
\label{testing}
\end{table}

\section{CONCLUSION}
In this paper, we have explored how RGB images can assist 3D vehicle detection and proposed an end-to-end single-stage feature adaptive fusion network by extending the recently proposed PointPillars, to achieve fast speed and effectively eliminate false positive. Based on the channel attention mechanism, we propose two fusion techniques: \textit{PointAttentionFusion} uses the channel attention mechanism to perform point-wise feature fusion of the two modalities; \textit{DointAttentionFusion} converts the image and point cloud into three modalities, and then performs pillar-wise feature fusion of multi-modalities. Researchers can choose different fusion methods according to actual needs. Evaluation on the KITTI dataset demonstrates significant improvement in filtering false positive over the approach that uses only a single modality. Furthermore, the proposed method yields competitive results and has the fastest speed compared to the published state-of-the-art multi-modal methods in the KITTI benchmark. In the future, we plan to explore how images can bring more performance enhancements to 3D vehicle detection and the gains images can bring to multi-class detection networks.


%%%%%%%%%%%%%%%%%%%%%%%%%%%%%%%%%%%%%%%%%%%%%%%%%%%%%%%%%%%%%%%%%%%%%%%%%%%%%%%%

%%%%%%%%%%%%%%%%%%%%%%%%%%%%%%%%%%%%%%%%%%%%%%%%%%%%%%%%%%%%%%%%%%%%%%%%%%%%%%%%

%%%%%%%%%%%%%%%%%%%%%%%%%%%%%%%%%%%%%%%%%%%%%%%%%%%%%%%%%%%%%%%%%%%%%%%%%%%%%%%%

%\section*{ACKNOWLEDGMENT}

%%%%%%%%%%%%%%%%%%%%%%%%%%%%%%%%%%%%%%%%%%%%%%%%%%%%%%%%%%%%%%%%%%%%%%%%%%%%%%%%
	
\bibliographystyle{IEEEtran}
\bibliography{IEEEabrv,bibtex/multi_view_representation}

\end{document}